\documentclass{article}
\usepackage{ijcai25}

\usepackage{times}
\usepackage{soul}
\usepackage{url}
\usepackage[hidelinks]{hyperref}
\usepackage[utf8]{inputenc}
\usepackage[small]{caption}
\usepackage{graphicx}
\usepackage{amsmath}
\usepackage{amsthm}
\usepackage{booktabs}
\usepackage{algorithm}
\usepackage{algorithmic}
\usepackage[switch]{lineno}
\usepackage{bm}
\usepackage{multirow}
\usepackage{amssymb}
\usepackage{subfig}

\usepackage{xcolor}

\title{Modality-Guided Dynamic Graph Fusion and Temporal Diffusion for Self-Supervised RGB-T Tracking}

\author{
Shenglan Li$^{1,2}$,
Rui Yao$^{1,2,}$\thanks{Corresponding author.},
Yong Zhou$^{1,2}$,
Hancheng Zhu$^{1,2}$,\\
Kunyang Sun$^{1,2}$,
Bing Liu$^{1,2}$,
Zhiwen Shao$^{1,2}$,
Jiaqi Zhao$^{1,2}$\\
\affiliations
$^1$School of Computer Sciences and Technology, China University of Mining and Technology\\
$^2$Mine Digitization Engineering Research Center of the Ministry of Education, China\\
\emails
\{shenglanli, ruiyao, yzhou, zhuhancheng, kunyang\_sun, liubing, zhiwen\_shao\}@cumt.edu.cn
}

\begin{document}

\maketitle

\begin{abstract}
To reduce the reliance on large-scale annotations, self-supervised RGB-T tracking approaches have garnered significant attention. However, the omission of the object region by erroneous pseudo-label or the introduction of background noise affects the efficiency of modality fusion, while pseudo-label noise triggered by similar object noise can further affect the tracking performance. In this paper, we propose GDSTrack, a novel approach that introduces dynamic graph fusion and temporal diffusion to address the above challenges in self-supervised RGB-T tracking. GDSTrack dynamically fuses the modalities of neighboring frames, treats them as distractor noise, and leverages the denoising capability of a generative model. Specifically, by constructing an adjacency matrix via an Adjacency Matrix Generator (AMG), the proposed Modality-guided Dynamic Graph Fusion (MDGF) module uses a dynamic adjacency matrix to guide graph attention, focusing on and fusing the object’s coherent regions. Temporal Graph-Informed Diffusion (TGID) models MDGF features from neighboring frames as interference, and thus improving robustness against similar-object noise. Extensive experiments conducted on four public RGB-T tracking datasets demonstrate that GDSTrack outperforms the existing state-of-the-art methods. 
The source code is available at \url{https://github.com/LiShenglana/GDSTrack}.
\end{abstract}

\section{Introduction}
The RGB-T tracking task, which leverages the complementary strengths of RGB images (rich texture information) and thermal infrared images (enhanced nighttime perception), has gained increasing attention in recent years~\cite{MambaVT,STTrack}. However, the reliance on manual annotation for both modalities is labor-intensive and resource-demanding~\cite{test-timeAdaption}, and the inherent inconsistency in cross-modal annotation further complicates the creation of high-quality RGB-T datasets. Consequently, the development of robust self-supervised methods for RGB-T tracking~\cite{S2OTFormer} has become crucial and offers significant practical advantages. 

To fully leverage information in images~\cite{Visual-Perception}, a class of mainstream methods relies on pseudo-labels \cite{USOT,ULAST,Diff-Tracker} for supervision.
USOT~\cite{USOT} relies on unsupervised optical flow to detect moving objects and applies a dynamic programming algorithm to establish interframe associations, thereby generating pseudo-labels to provide supervision. 
\begin{figure}[t]
\centering
\includegraphics[width=1\linewidth]{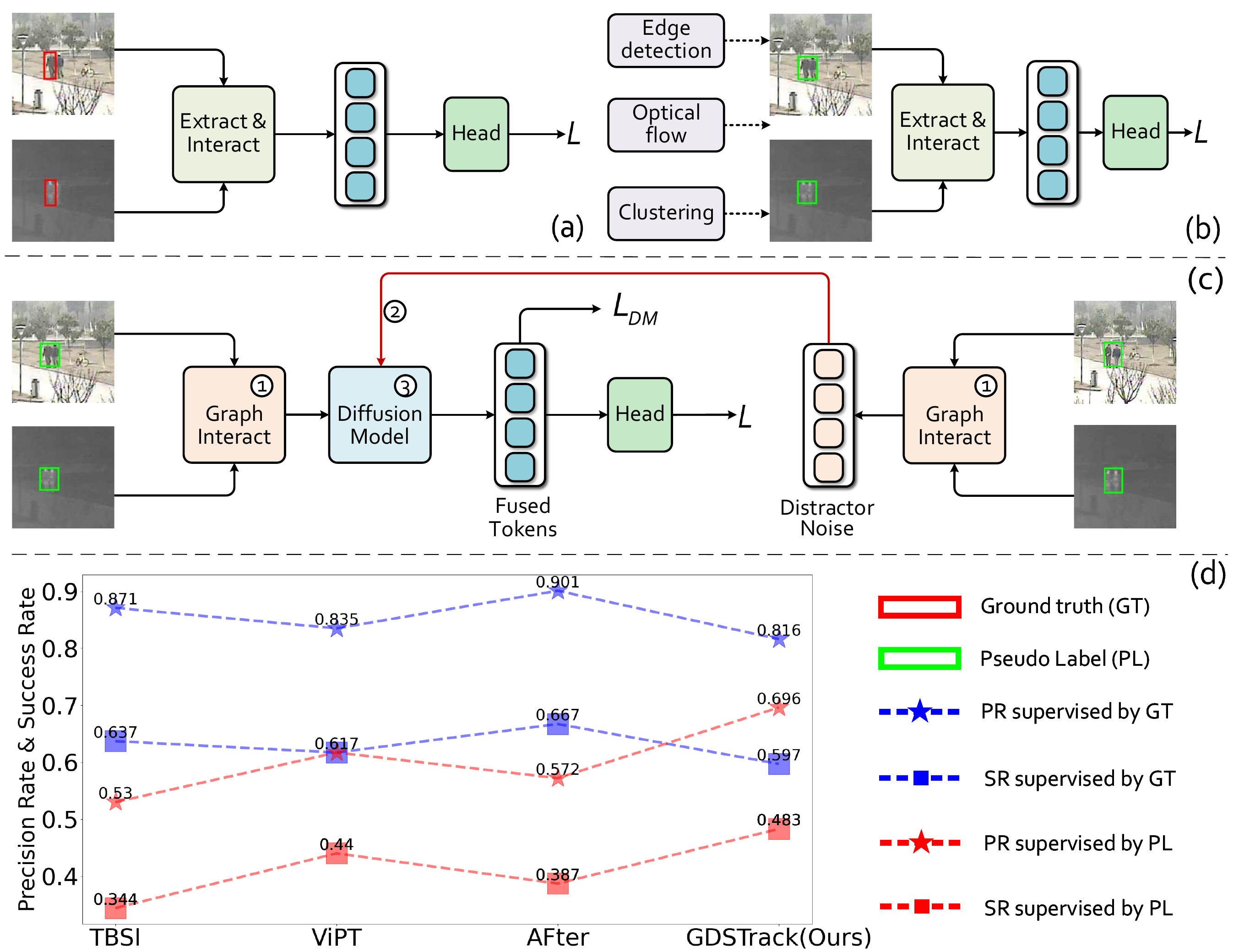}   \caption{(a) Existing RGB-T tracking methods rely heavily on a large number of ground truth annotations, making it challenging to handle larger-scale RGB-T datasets under supervised settings. (b) The discrepancy between the pseudo label and ground truth makes accurate object tracking difficult, causing performance degradation when used for the fully supervised training of sota trackers. (c) We leverage modality-guided dynamic graph fusion and temporal diffusion to address the issues of object irrelevant fusion and distractor noise caused during the training process supervised by pseudo-label. (d) Compared with the three trackers, our proposed method achieved the best performance under pseudo-label supervision.}
\label{intro}
\end{figure}
However, a discrepancy between the pseudo-labels and ground truth is inevitable. For example, comparing the accurate red bounding box in Fig.\ref{intro}(a) with the pseudo green bounding box in Fig.\ref{intro}(b), it is evident that the pseudo-label incorrectly includes both the object and a similar moving object, thereby further exacerbating performance degradation. As illustrated in Fig.~\ref{intro}(d), when training with pseudo-labels, a significant performance degradation is observed in TBSI~\cite{TBSI}, ViPT~\cite{ViPT}, and AFter~\cite{after}. Obviously, this performance degradation stems from pseudo-label noise, which arises from two primary factors: (1) inaccurate pseudo-labels may fail to capture whole object regions and always introduce excessive background noise, and (2) inaccurate annotation caused by interference from similar objects will significantly degrade the model performance. To address these two problems, we designed a self-supervised RGB-T tracking method as shown in Fig.~\ref{intro}(c). 

In fact, label noise also exists in fully supervised tasks, where misalignment between modality labels can introduce label noise. GMMT~\cite{GMMT} models noise as Gaussian noise, and uses a generative model~\cite{DDIM,diffuseforVAD} to enhance its ability to perceive noise. However, when dealing with pseudo-labels, simple Gaussian noise is insufficient for modeling the two types of labeling errors mentioned above. To address the pseudo-label noise issue, we propose our method. Specifically, to tackle the first issue, we introduce the Modality-Guided Dynamic Graph Fusion (MDGF) module, which dynamically adjusts the adjacency matrix based on the similarity between the input RGB and thermal infrared modalities. By leveraging a dynamic adjacency matrix to guide graph attention in mining and focusing on object regions, a fusion that supports tracking is achieved, even in the presence of inaccurate annotations. Moreover, to study the second issue, we draw inspiration from GMMT~\cite{GMMT} and propose a Temporal Graph-Informed Diffusion (TGID) module, which treats fused information from neighboring frames as noise within a diffusion model. This approach helps the model to identify interference from similar objects and train it to be more robust to such noise. Furthermore, the MDGF module complements the TGID module to prevent information loss and enhance overall model performance.  In summary, our main contributions are as follows:
\begin{itemize}
\item We propose a novel self-supervised RGB-T tracker GDSTrack, which leverages multi-modal optical flow to extract pseudo-labels for coarse localization, significantly reducing the cost of manual annotation.     \item We design the MDGF module to simulate interference from similar objects in pseudo-labels and utilize the generative capability of the TGID module for denoising and fine localization, achieving a balance between performance and cost.
\item Ablation studies and comparative experiments on four benchmarks demonstrate the state-of-the-art performance of our method and its ability to enhance self-supervised learning.
\end{itemize}
\section{Related Work}
\noindent \textbf{Self-supervised Tracking.} RGB-T tracking has advanced significantly owing to the complementary nature of RGB and thermal infrared modalities. \cite{MambaVT} leverage spatio-temporal contextual modeling through long-range cross-frame integration and short-term historical trajectory prompts. \cite{STTrack} introduce a temporal state generator that produces temporal information tokens to guide the localization of the object in the next time state. Self-supervised RGB-T tracking~\cite{S2OTFormer} addresses the challenge of requiring large-scale manual annotations for tracking tasks. However, this topic has not received sufficient research attention. Most self-supervised tracking methods rely primarily on the RGB modality alone. The UDT~\cite{UDT} exploits the discriminative power of correlation filtering for tracking and employs cycle consistency as a self-supervised signal. Existing deep trackers can be trained using synthesized data in routine ways, without requiring human annotation.  \cite{S2Siamfc} exploit the fact that an image and any cropped region of it naturally form a pair for self-training. To replace naive cropping methods such as center cropping, \cite{USOT} propose a more accurate pseudo-label generation method based on optical flow and dynamic programming techniques. Building on this, \cite{Diff-Tracker} learn a prompt representation of an object using a pre-trained diffusion model. However, these methods do not account for interference caused by similar objects in the absence of ground-truth labels. Our method tackles the root cause of the noise by simulating distractor noise using the modal fusion results of similar objects, thereby enhancing the robustness of the model to such interference.

\noindent \textbf{RGB-T Fusion} aims to fully utilize the complementary characteristics of RGB and thermal infrared images to enhance feature extraction and improve performance in downstream tasks.  In terms of the fusion level, \cite{multi-fusion} consider several fusion mechanisms at the pixel, feature, and response levels.  In terms of fusion methods, \cite{CBPNet} use a channel attention mechanism to implement the adaptive calibration of feature channels before realizing hierarchical feature fusion. \cite{DATFuse} propose a novel model for infrared and visible image fusion via a dual attention Transformer to represent long-range context information. \cite{GMMT} seek to uncover the potential of generative techniques to address the critical challenge in multi-modal tracking. \cite{SeqTrack} predict bounding boxes through a sequence-to-sequence framework. \cite{ViPT} introduce prompt learning for multi-modal fusion tracking method. In terms of frequency domain fusion, \cite{CDDFuse} decompose modality-specific and modality-shared features to better reflect the semantic information contained in the high-frequency and low-frequency features. However, these methods do not account for the fusion process and focus on irrelevant areas because of the lack of ground truth labels. Our method can dynamically generate adjacency matrices, guiding the graph attention model to focus more on object regions.
\begin{figure*}[ht]
  \centering
  \includegraphics[width=\linewidth]{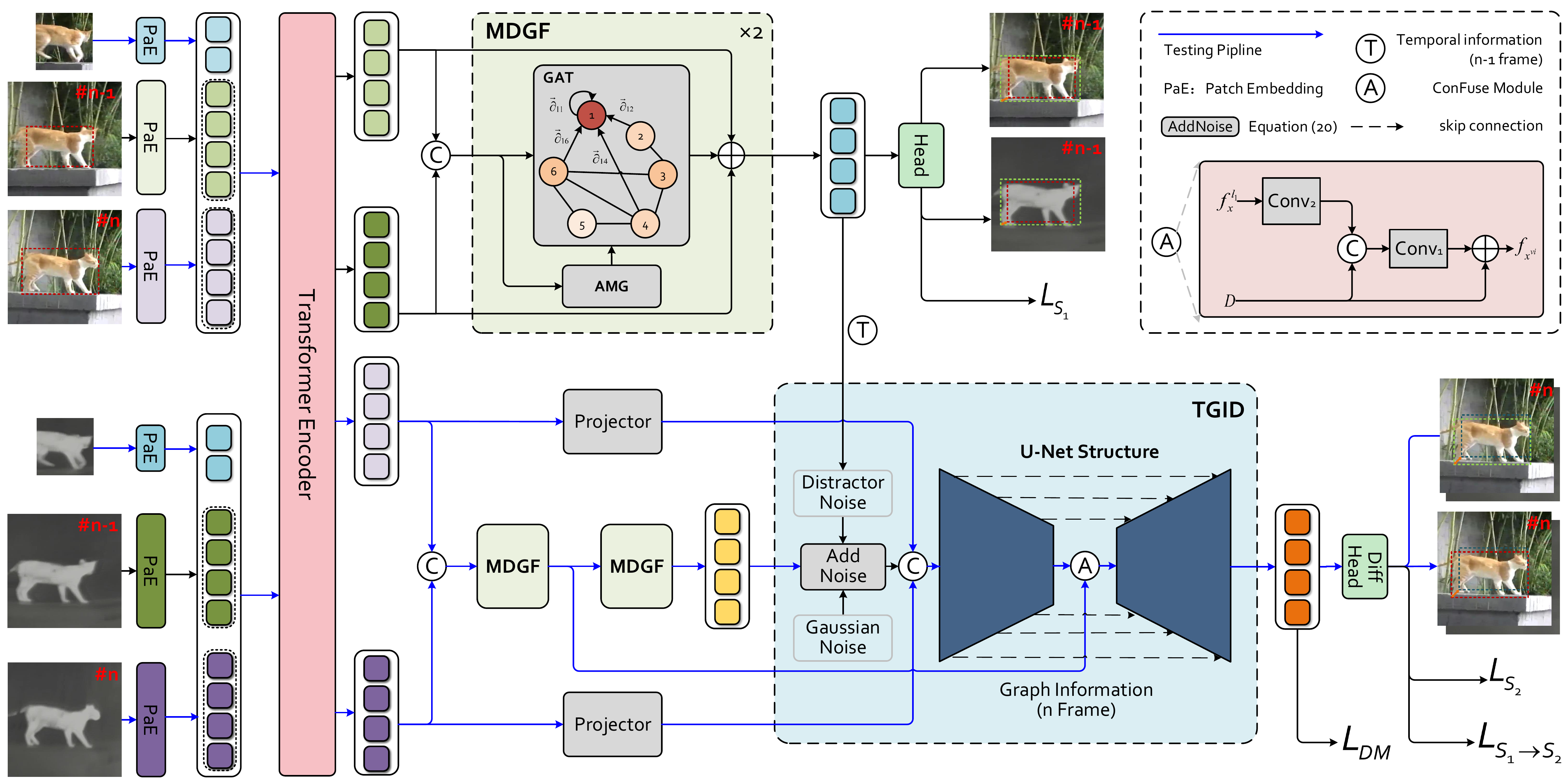}
  \caption{The pipeline of GDSTrack model. First, the encoder obtains the search frame features that interact with the template frame features. Subsequently, the MDGF module fuses the features of the two modalities using graph attention guided by dynamic adjacency matrix. Then the features obtained from the MDGF module are used as distractor noise and input for the TGID module to perform denoising and obtain the final tracking results. We use TGID to enhance the model's robustness to noise.}
  \label{pipeline}
\end{figure*}
\section{Methodology}
\subsection{Framework Overview} Our model performs self-supervised tracking by fully leveraging the advantages of both RGB and infrared features through the Modality-Guided Dynamic Graph Fusion and Temporal Graph-Informed Diffusion modules, as shown in Fig.~\ref{pipeline}. GDSTrack obtains the interaction features through the encoder. The RGB and infrared images share encoder parameters. We first use the AMG module to dynamically adjust the adjacency matrix based on the features of the two modalities, guiding graph attention in MDGF to fuse the information from both modalities. The TGID module then denoises similar object noise simulated by the MDGF module. Finally, the output is fed into the tracking head for tracking.  GDSTrack framework takes two RGB images and their corresponding infrared images as input: the visible template frame $\bm{z}^v\in\mathbb{R}^{3\times W_{z_0} \times H_{z_0}}$ cropped from the first visible frame $\bm{x}_1^v\in\mathbb{R}^{3\times W_{x_0} \times H_{x_0}}$, and the infrared template frame $\bm{z}^i\in\mathbb{R}^{1\times W_{z_0} \times H_{z_0}}$ cropped from the first infrared frame $\bm{x}_1^i\in\mathbb{R}^{1\times W_{x_0} \times H_{x_0}}$. The second visible search frame $\bm{x}_2^v\in\mathbb{R}^{3\times W_{x_0} \times H_{x0}}$, the second infrared search frame $\bm{x}_2^i\in\mathbb{R}^{1\times W_{x_0} \times H_{x_0}}$, where $W_{z_0}$ and $H_{z_0}$ are half of $W_{x_0}$ and $H_{x_0}$, respectively. We obtain the RGB and infrared search frame features, denoted as $\bm{f}_{x_1}^v\in\mathbb{R}^{W_xH_x \times d_{model}}$ and $\bm{f}_{x_1}^i\in\mathbb{R}^{W_xH_x \times d_{model}}$, respectively, after the interaction between the template and the search frame through the encoder. The MDGF module performs an initial fusion of the inputs $\bm{f}_{x_1}^v\in\mathbb{R}^{W_xH_x \times d_{model}}$ and $\bm{f}_{x_1}^i\in\mathbb{R}^{W_xH_x \times d_{model}}$, producing two layers of fused results $\bm{f}_{x_1}^{l_1}$ and $\bm{f}_{x_1}^{S_1}$. \begin{align}
\bm{f}_{x_1}^{l_1}, \bm{f}_{x_1}^{S_1} &= MDGF(\bm{f}_{x_1}^v, \bm{f}_{x_1}^i), 
\end{align}
\noindent where $MDGF(\cdot)$ indicates the Modality-guided Adaptive Graph Fusion Module (see Sect.~\ref{MDGF}). $\bm{f}_{x_1}^{S_1}$ is processed through the tracking head to obtain the tracking score $Score^{S_1}$ and the predicted bounding box $B^{S_1}$: 
\begin{align}
Score^{S_1}, B^{S_1} &= Head(\bm{f}_{x_1}^{S_1}), 
\end{align} 
\noindent where we use a series of Conv-BN-ReLU layers to independently estimate the top-left and bottom-right corners following~\cite{mixformer}. We employ the Generalized Intersection over Union (GIoU) loss and $L_1$ loss to guide tracker training. The loss function in stage one is defined as follows:
\begin{align}
L_{S_1} &= \lambda_1L_{GIoU}(B^{S_1}, \hat{B}) + \lambda_2L_1(B^{S_1}, \hat{B}). 
\end{align}
\noindent Where $\hat{B}$ is our pseudo label, and we set the loss weight $\lambda_1$ to 2 and set $\lambda_2$ to 5 following~\cite{MAT_2023_CVPR}. The third search frame is then introduced to obtain the corresponding visible and thermal infrared features, named as $\bm{f}_{x_2}^v\in\mathbb{R}^{W_xH_x \times d_{model}}$ and $\bm{f}_{x_2}^i\in\mathbb{R}^{W_xH_x \times d_{model}}$. 

We first use the MDGF module to obtain the first-layer and second-layer fusion results, denoted as $\bm{f}_{x_2}^{l_1}$ and $\bm{f}_{x_2}^{S_1}$:
\begin{align}
\bm{f}_{x_2}^{l_1}, \bm{f}_{x_2}^{S_1} &= MDGF(\bm{f}_{x_2}^v, \bm{f}_{x_2}^i).
\end{align}
We then utilize the TGID module to further denoise $\bm{f}_{x_2}^{S_1}$ from the MDGF to obtain more refined fused features $\bm{f}_{x_2}^{S_2}$:
\begin{align}
\bm{f}_{x_2}^{S_2} &= TGID(\bm{f}_{x_1}^{S_1}, \bm{f}_{x_2}^{l_1}, \bm{f}_{x_2}^{S_1}, \bm{f}_{x_2}^v, \bm{f}_{x_2}^i), 
\end{align}
\noindent where $TGID(\cdot)$ indicates the Temporal Graph-Informed Diffusion Module (see Sect.~\ref{TGID}). We use $\bm{f}_{x_2}^v$ and $\bm{f}_{x_2}^i$ as the initial conditions for the diffusion model, with $\bm{f}_{x_2}^{l_1}$ from the first layer of MDGF serving as the conditions for the intermediate layers of the denoising process. $\bm{f}_{x_1}^{S_1}$ is set as distractor noise, contributing to the noise during the added noise process. $\bm{f}_{x_2}^{S_2}$ is used to obtain the tracking results through DiffHead:
\begin{align}
Score^{S_2}, B^{S_2} &= DiffHead(\bm{f}_{x_2}^{S_2}),
\end{align}
\noindent where DiffHead shares the same structure as Head in MDGF.  Then, we employ pseudo-labels $\hat{B}$ to supervise the training: \begin{align}
L_{S_2} &= \lambda_1L_{GIoU}(B^{S_2}, \hat{B}) + \lambda_2L_1(B^{S_2}, \hat{B}).
\end{align}
We use the predicted bounding boxes $\hat{B}^{S_1}$ from the MDGF to supervise the training process of the diffusion model: 
\begin{align}
L_{S_1\rightarrow S_2} &= \lambda_1L_{GIoU}(B^{S_2}, \hat{B}^{S_1}) + \lambda_2L_1(B^{S_2}, \hat{B}^{S_1}).
\end{align}
The final loss function in stage two is defined as follows: \begin{align}
L &= L_{S_2} + L_{S_1\rightarrow S_2} + \lambda L_{DM},
\end{align}
\noindent where $L_{DM}$ is the generative loss.

\subsection{Modality-guided Dynamic Graph Fusion}
\label{MDGF}
\begin{figure}[t]
  \centering
  \includegraphics[width=1\linewidth]{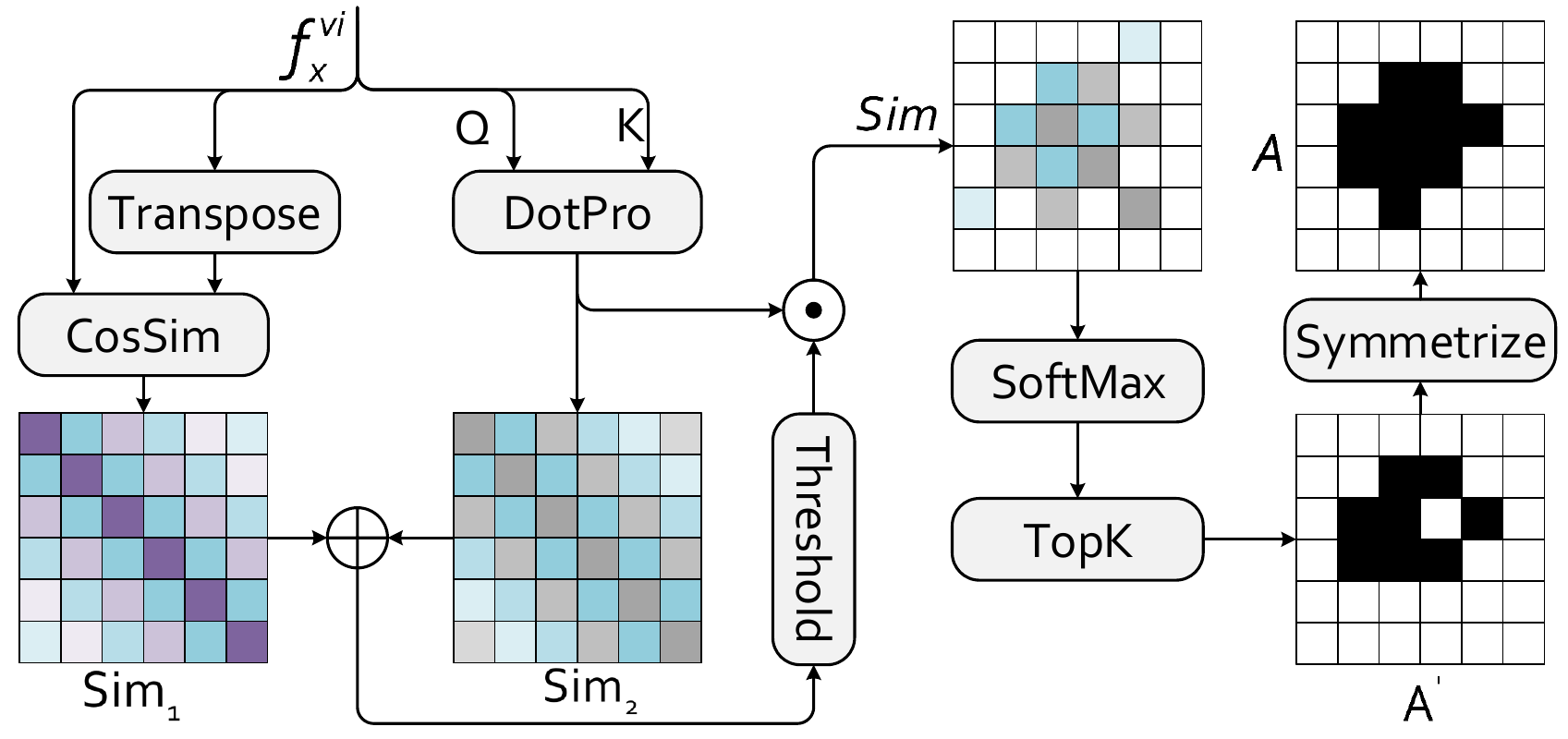}
  \caption{The pipeline of AMG. The AMG module concatenates visible and infrared features along the sequence dimension, then computes cosine similarity and dot-product attention with themselves. After summation and threshold filtering, they guide the dot-product attention to generate similarity mask. Finally, SoftMax, TopK, and Symmetrize are applied to obtain the adjacency matrix.}
  \label{AMG}
\end{figure}
The MDGF consists of an Adjacency Matrix Generator and a Graph Attention Network. AMG generates a dynamic adjacency matrix based on the similarity between RGB and infrared modalities to guide the graph attention network. This similarity-based fusion method enables GDSTrack to focus more on the coherent regions of the object, even in the absence of ground truth labels.

\noindent \textbf{Adjacency Matrix Generator.} The AMG module employs Cosine Similarity to create a mask that guides the cross-attention mechanism in generating the corresponding adjacency matrix according to feature variations (see Fig.~\ref{AMG}). 

We concatenate $\bm{f}_{x}^v$ and $\bm{f}_{x}^i$ along the sequence dimension:
\begin{align}
\bm{f}_{x}^{vi} &= Concat(\bm{f}_{x}^v, \bm{f}_{x}^i),
\end{align}
\noindent where we get $\bm{f}_{x}^{vi}\in\mathbb{R}^{2W_xH_x \times d_{model}}$. Then, we compute the similarity $Sim_1$ between $\bm{f}_{x}^v$ and $\bm{f}_{x}^i$ using the scaled dot product attention ~\cite{Transformer}:
\begin{align}
Q = \bm{f}_{x}^{vi}W^Q,
K = \bm{f}_{x}^{vi}W^K,
S_1 = \frac{Q K^{\mathrm{T}}}{\sqrt{d_k}},
\end{align}
\noindent where the projections are the parameter matrices $W^Q\in\mathbb{R}^{d_{model}\times d_k}$ and $W^K\in\mathbb{R}^{d_{model} \times d_k}$. $\frac{1}{\sqrt{d_k}}$is a scaling factor and $d_k$ is 2048. We then calculate the Cosine Similarity $Sim_2$ between $\bm{f}_{x}^v$ and $\bm{f}_{x}^i$:
\begin{align}
S_2 &= CosineSim(\bm{f}_{x^{vi}}, \bm{f}_{x^{vi}}^{\mathrm{T}}).
\end{align}

We sum the two similarities and then shift their values to the range $[0, 1]$:
\begin{align}
M &= (S_1 + S_2 + 1) / 2,
\end{align}
\noindent where $M\in\mathbb{R}^{2W_xH_x \times 2W_xH_x}$.

We use $M$ as a mask, setting the similarity values to less than the threshold $\theta$ to negative infinity, and retaining the similarity values from $Sim_1$ otherwise:
\begin{equation}
\left\{
\begin{aligned}
S_{i,j} &= S_{1_{i,j}}, if M > \theta,\\
S_{i,j} &= -\infty, if M \leq \theta.\\
\end{aligned}
\right.
\end{equation}

\noindent Then we get the final similarity:
\begin{align}
S' &= softmax(S).
\end{align}

\noindent We then set the Top-K positions with the largest similarity values in the final similarity matrix to 1, while setting all other positions to zero:
\begin{equation}
\left\{
\begin{aligned}
A'_{i,j} &= 1, if\ S'_{i,j} \geq \text{the }k\text{-th largest values of }S',\\
A'_{i,j} &= 0, otherwise.\\
\end{aligned}
\right.
\end{equation}

\noindent To ensure the symmetry of adjacency $A'$:
\begin{align}
A = (A' + A'^{\mathrm{T}}) / 2,
\end{align}
\noindent where $A\in\mathbb{R}^{2W_xH_x \times 2W_xH_x}$.

\noindent \textbf{Graph Attention Networks.}
We utilize a two-layer Graph Attention Network (GAT)~\cite{GAT} to perform graph-based attention fusion of visible and thermal infrared features under the guidance of the adjacency matrix generated by the AMG module:
\begin{align}
\bm{f}_{x}^{l_1}, \bm{f}_{x}^{S_1} &= GAT(\bm{f}_{x}^{vi}, A).
\end{align}
GAT takes each pixel in the concatenated modalities $\bm{f}_{x}^{vi}$ as a graph node, and the adjacency matrix $A$ preserves both intra-modal and inter-modal relationships. The input features are processed using weighted learning, where the attention mechanism adaptively assigns weights based on the correlations between features. 

In the first layer, a learnable matrix $W_1\in\mathbb{R}^{d_{model} \times nhid}$ is used to perform a linear transformation of features $\bm{f}_{x}^{vi}$. Then, a shared attentional mechanism and softmax function is performed following the GAT model. Similarly, a learnable vector $W_2\in\mathbb{R}^{nhid \times out_{model}}$ is used to perform a linear transformation of the output of the first layer, $\bm{f}_{x}^{layer1}$.

To reduce information loss, the output of the second layer $\bm{f}_{x}^{l2}$ is added to the visible and infrared features after the dimension reduction using a $1 \times 1$ convolution as $Proj(\cdot)$:
\begin{align}
\bm{f}_{x}^{S_1} &= Add(split(\bm{f}_{x}^{l_2})) + Proj(\bm{f}_{x}^v + \bm{f}_{x}^i),
\end{align}
\noindent where $split$ is the inverse of Concatenation.

\subsection{Temporal Graph-Informed Diffusion}
\label{TGID}
To address the noise from similar objects during pseudo-label training, we designed the TGID module. It treats the MDGF results of nearby frames as noise using the first-layer MDGF features to improve robustness to noise and prevent information loss. Specifically, after obtaining the fusion results $\bm{f}_{x_2}^{S_1}$ from MDGF, we introduce a diffusion model DDIM~\cite{DDIM} to denoise $\bm{f}_{x_2}^{S_1}$. By leveraging the generative capabilities of the diffusion model, we achieved fusion results that are more conducive to tracking. We use $\bm{f}_{x_2}^{S_1}$ as the input and train the model with the features of the two modalities $\bm{f}_{x_2}^v$ and $\bm{f}_{x_2}^i$ as conditions following~\cite{GMMT}, where we use a $1\times1$ convolution as a projector for feature dimensionality reduction. We also use the first-layer output of MDGF $\bm{f}_{x_2}^{l_2}$ as another condition. Additionally, distractor object noise $\bm{f}_{x_1}^{S_1}$ is introduced during the diffusion process to enhance the robustness of the model against similar distractor.

\noindent \textbf{Diffusion process.} Due to the motion in video frames, the fusion result of the neighboring frame $\bm{f}_{x_1}^{S_1}$ is offset at the object location compared to the fusion result of the current frame $\bm{f}_{x_2}^{S_1}$. This natural offset can serve as a similarity distractor for the current frame, enhancing the robustness of the model. Based on this observation, in addition to the original Gaussian noise, we incorporate the first-stage fusion result $\bm{f}_{x_1}^{S_1}$ as part of the noise in the diffusion process. 

In the forward diffusion process, we use the MDGF fusion output $\bm{f}_{x_2}^{S_1}$ as $x_0$. Then $x_0$ undergoes diffusion through the random Gaussian noise $\bar{z}_t$ and the distractor noise from the first-stage fusion result of the neighboring frame $\bm{f}_{x}^{S_1}$ as $d_t$:
\begin{align}
x_t &= \sqrt{\bar{\alpha}_t}x_0 + ( 1- \beta)\sqrt{1 - \bar{\alpha}_t}\bar{z}_t + \beta d_t.
\end{align}
\noindent \textbf{Denoising process.} Some recent works~\cite{diffuseforVAD} utilize additional information as conditions for generative models to guide the denoising process of the model. We employ a UNet~\cite{UNet} network to perform the denoising process in the model following~\cite{GMMT}. We design a simple module named the Condition Fuse (ConFuse) model to serve as supplementary information for the intermediate layers of the UNet network. We use the first-layer graph convolution results $\bm{f}_{x}^{l_1}$ generated in MDGF as middle conditions in the denoising process:
\begin{align}
\bm{f}_{x^{vi}} &= ConFuse(\bm{f}_{x}^{l_1}, D)\\
&= Conv_1(concat(Conv_2(\bm{f}_{x}^{l_1}), D)) + D.
\end{align}
\noindent The features are downsampled using $Conv_2(\cdot)$ to match the size of the intermediate layers of the UNet network named $D$, then concatenated with $D$ along the channel dimension. Subsequently, $Conv_1(\cdot)$ is used to reduce the dimensionality of the features, and finally, they are added to $D$ to prevent information loss.

We calculate the $L_2$ loss between noise and the output of the denoising process:
\begin{align}
L_{DM} &= L_2(noise, output).
\end{align}
\section{Experiments}

\textbf{Datasets.} \textbf{GTOT} dataset~\cite{GTOT} comprises 50 pairs of visible and thermal infrared video sequences and corresponding ground truth annotations. These sequences are designed to examine diversity and biases across seven specific challenges. Attribute-based annotations allow for a more detailed evaluation of tracker performance.  
\textbf{RGBT234}~\cite{RGBT234} encompassing 234 sequences, with the longest sequence containing up to 8k frames. This dataset is annotated with 12 attributes, making it a robust benchmark for assessing the effectiveness of different trackers.  
\textbf{LasHeR} dataset~\cite{LasHeR} featuring 1,224 pairs of visible and thermal infrared videos, with a total of 730k frame pairs. It introduces 19 real-world challenge attributes, providing a more extensive evaluation framework than GTOT and RGBT234.  
\textbf{VTUAV}~\cite{VTUAV} is a large-scale benchmark specifically designed for visible-thermal UAV tracking, including 500 sequences and 1.7 million high-resolution frame pairs. Its diversity encompasses various categories and scenes, supporting exhaustive evaluations that cover short-term tracking, long-term tracking, and segmentation mask prediction.

\noindent \textbf{Metric.} Following~\cite{MFGNet,after}, we evaluate our tracker using three metrics: \textbf{Precision rate} (PR), \textbf{Normalized precision rate} (NPR), and \textbf{Success rate} (SR). PR measures the proportion of frames where the Euclidean distance between the predicted object's center and the ground truth center is below a predefined threshold. Specifically, we use a threshold of 5 pixels for GTOT and 20 pixels for RGBT234 and LasHeR. PR is further normalized as NPR.
SR assesses the intersection ratio between the predicted bounding box and the ground truth bounding box, also referred as the overlap ratio. It quantifies the percentage of frames where the overlap exceeds a certain threshold. For all four datasets, we calculate the area under the success rate curve (AUC) to derive the success score.

\begin{table}[t]
\centering
\resizebox{.5\textwidth}{!}{
\begin{tabular}{c | c c | c c  c |c c | c c} 
 \toprule
 \multirow{2}{*}{Tracker} &  \multicolumn{2}{c|}{RGBT234} & \multicolumn{3}{c|}{LasHeR}& \multicolumn{2}{c|}{VTUAV}&\multicolumn{2}{c}{GTOT}\\
      & PR$\uparrow$ & SR$\uparrow$ & PR$\uparrow$ & NPR$\uparrow$ & SR$\uparrow$ & PR$\uparrow$ & SR$\uparrow$& PR$\uparrow$ & SR$\uparrow$\\ 
   \midrule
    KCF\(\dagger\)& 46.3 & 30.5 & - & - & - & - & - & - & - \\
    MEEM\(\dagger\) & 63.6 & 40.5 & - & - & - & - & - & 64.8 & 52.3\\ 
    UDT\(\dagger\)& 56.8 & 42.0 & - & - & - & - & -& 73.7 & 61.4 \\
    USOT\(\dagger\) & 50.8 & 31.8 & 24.8 & 20.6 & 16.0 & 36.3 & 26.5& 70.6 & 58.3 \\
    UDT-FF & 56.1 & 41.4 & 28.1 & 22.8 & 21.5 & 51.0 & 40.1&  79.3 & 65.1 \\
    TBSI$^\ast$ & 53.0 & 34.4 & 29.8 & 24.5 & 25.9 & 24.4 & 23.4& 57.2 & 49.1\\
    ViPT$^\ast$ &  61.7 & 44.0 & 38.2 & 34.1 & 32.4 & 47.8 & 42.4&61.1 & 54.0\\
    AFter$^\ast$& 57.2 & 38.7 & 30.7 & 25.9 & 26.6 & 25.1 & 22.1& 30.7 & 25.7\\
    S2OTFormer & 68.4 & 47.7 & 39.8 & 35.4 & 29.5 & 56.7 & 44.5& $\bm{83.1}$ & $\bm{70.2}$\\
    GDSTrack & $\bm{70.9}$ & $\bm{48.5}$ & $\bm{45.9}$ & $\bm{39.3}$ & $\bm{35.4}$ & $\bm{72.3}$ & $\bm{59.8}$& 73.9 & 59.8\\
 \bottomrule
\end{tabular}
}
\caption{PR, NPR, and SR evaluation results compared to state-of-the-art self-supervised methods on four benchmarks. Among them, KCF and MEEM are correlation filtering-based methods. Trackers marked with \(\dagger\) denote the results obtained by directly combining RGB and infrared features using existing self-supervised RGB tracking methods. For a fair comparison, trackers with $^\ast$ refer to the results trained with the same pseudo-labels as GDSTrack (Ours). Performance of which achieved best are marked in $\bm{bold}$.}
\label{sota-selfsupervised}
\end{table}
\begin{figure}[!ht]
\centering
\subfloat[]{\includegraphics[width=1.7in]{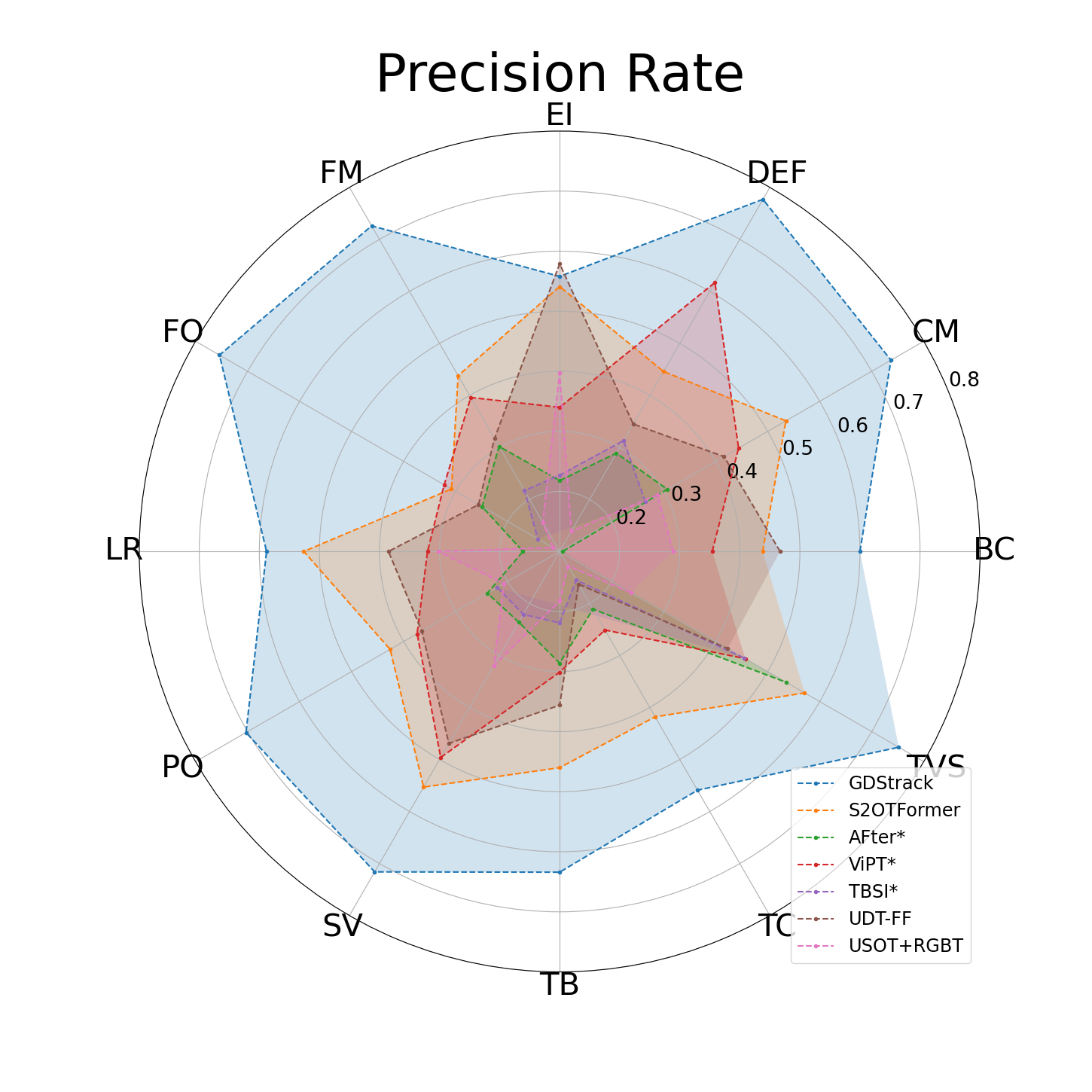}}%
\subfloat[]{\includegraphics[width=1.7in]{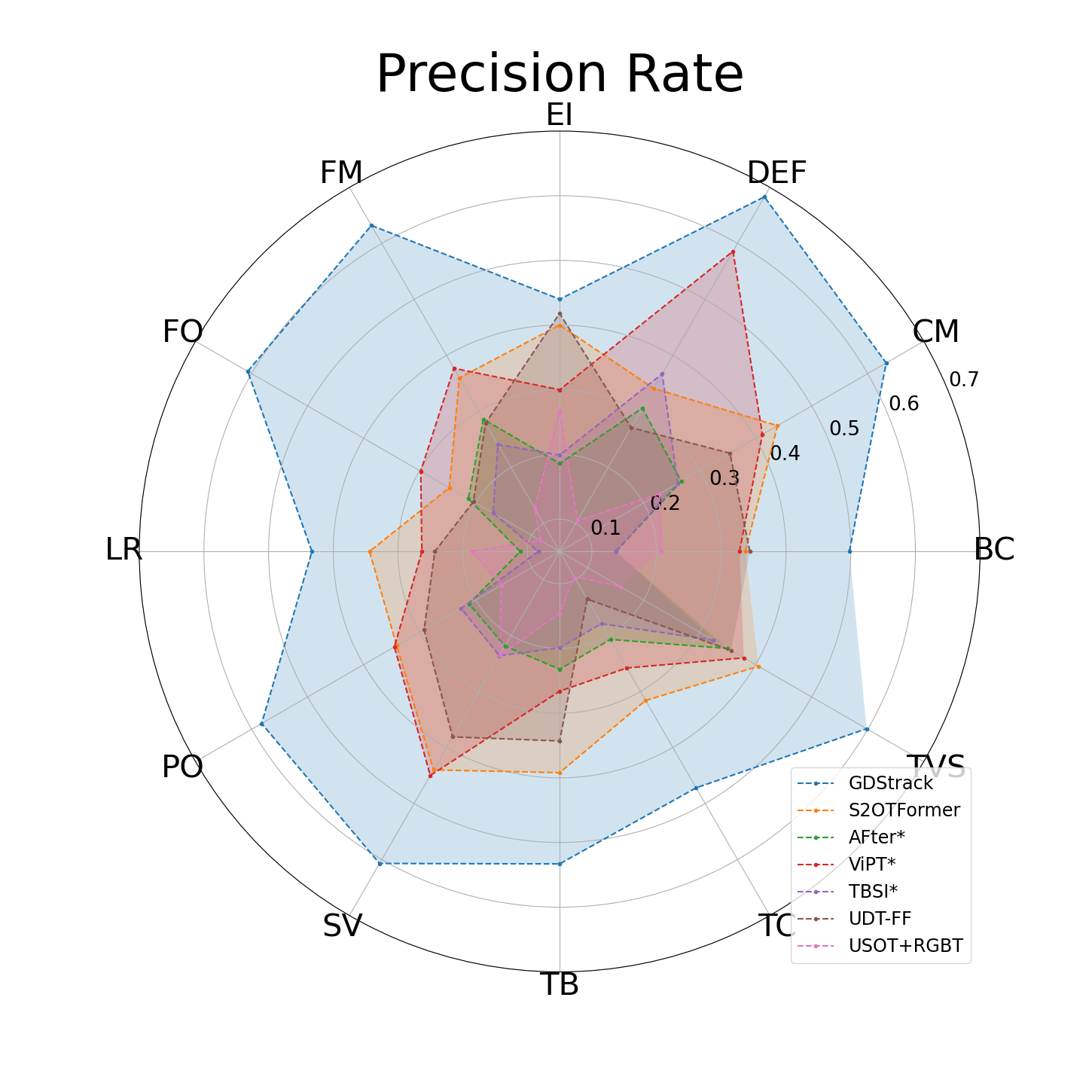}}%
\vspace{-3mm}
\caption{Attribute-based evaluation on VTUAV dataset compared against five self-supervised RGBT trackers. (a) Precision Rate with different attributes. (b) Success Rate with different attributes.}
\vspace{-5mm}
\label{VTUAV-attr}
\end{figure}

\begin{figure}[t]
\centering
\subfloat[]{\includegraphics[width=1.7in]{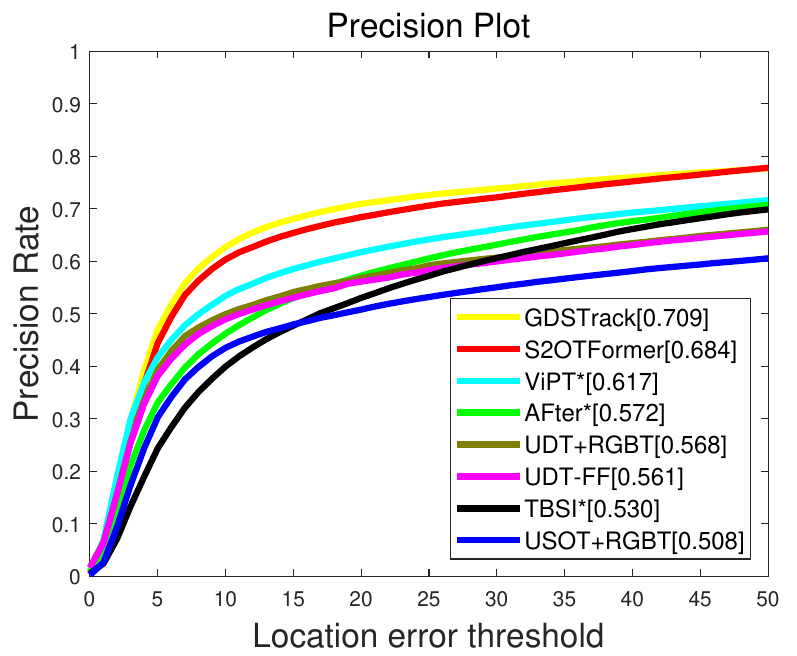}}%
\subfloat[]{\includegraphics[width=1.7in]{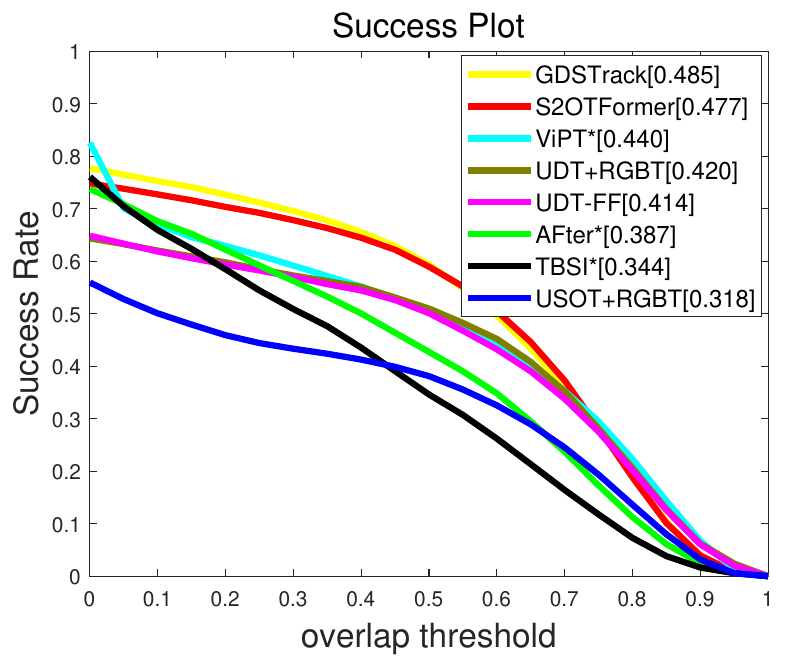}}%
\vspace{-3mm}
\caption{PR and SR on RGBT234 dataset compared against other self-supervised RGBT trackers.}
\label{RGBT234-PR-SR}
\end{figure}
\noindent \textbf{Implementation Details.} Our model is implemented using the PyTorch platform.
We use LasHeR~\cite{LasHeR} as our training dataset. Following~\cite{USOT}, we generate pseudo labels and confidence scores for both the visible and infrared datasets. The label with the higher confidence score is selected as the final pseudo label.
We use ViT-B/16~\cite{ViT} as the feature encoder for interaction between the template frame and the infrared frame. The encoder is initialized with the pre-trained parameters proposed in~\cite{MAT_2023_CVPR}. The template size and search region size are set to $128\times128$ and $256\times256$.
Training is carried out using the AdamW optimizer~\cite{AdamW}. The batch size is 16, and the learning rate is set to 0.00003. We first train the MDGF module and its corresponding tracking head. Starting from the 10th epoch, the parameters of the ViT encoder are unfrozen. From the 22nd epoch, the encoder, MDGF, and tracking head parameters are frozen, and only the parameters related to the diffusion model and the tracking head (diffuse-head) are trained. The training is guided by the MDGF tracking results and pseudo-labels. The training concludes at the 50th epoch. The weight decay coefficient is set to 0.0001, and the momentum value is set to 0.9.

\subsection{Main Results}
We present the main experimental results here, and additional results can be found in the appendix.

\noindent \textbf{Comparisons with the State-of-the-Art.} As shown in Tab. \ref{sota-selfsupervised}, we compared our method with other state-of-the-art self-supervised RGB-T tracking methods, including KCF~\cite{KCF}+ RGBT, MEEM~\cite{MEEM} + RGBT, TBSI$^\ast$~\cite{TBSI}, AFter$^\ast$~\cite{after}, ViPT$^\ast$~\cite{ViPT}, UDT~\cite{UDT} + RGBT, UDT-FF~\cite{UDT-FF}, USOT~\cite{USOT} + RGBT and S2OTFormer~\cite{S2OTFormer}. We re-trained AFter~\cite{after}, ViPT~\cite{ViPT} and TBSI~\cite{TBSI} on the LasHeR dataset using the same RGB-T pseudo-labels as our method, named as AFter$^\ast$, ViPT$^\ast$ and TBSI$^\ast$ for comparison.
It can be observed that our model achieves state-of-the-art performance on the RGBT234, LasHeR, and VTUAV datasets. 

It can be seen that our method exceeds the self-supervised method S2OTFormer (68.4\% / 47.7\%) by 2.5\% on PR and 0.8\% on SR on the RGBT234 dataset. It also surpasses the S2OTFormer (39.8\% / 35.4\% / 29.5\%) by 6.1\% on PR, 3.9\% on NPR, and 5.9\% on SR on the LasHeR dataset. It also surpasses the S2OTFormer (56.7\% / 44.5\%) by 15.6\% on the PR score and 15.3\% on the SR score on the VTUAV dataset. We hypothesize that the suboptimal performance on the GTOT dataset is due to the smaller number of sequences and shorter sequence lengths in this dataset. We compared our model with other state-of-the-art self-supervised RGB-T trackers on the RGBT234 dataset. As shown in Fig.~\ref{RGBT234-PR-SR}, our model achieved the best performance on both the PR plots and the SR plots.

\noindent \textbf{Attribute Analysis.} The performance of our method GDSTrack, along with other sota self-supervised methods on different attribute challenges of VTUAV dataset is shown in Fig.~\ref{VTUAV-attr}. There is no OV attribute in the Short-term Evaluation of the VTUAV dataset, so we only draw twelve attributes. As observed, our method achieves the best performance on most attributes of the VTUAV dataset with only 2.1\% of PR lower on extreme illumination (EI) attribute than UDT-FF.

\noindent \textbf{Speed Analysis.} As shown in Tab.~\ref{fps}, 
GDSTrack achieves better performance than S2OTFormer while maintaining comparable speed.
\begin{table}[t]
\centering
\resizebox{0.5\textwidth}{!}{
\begin{tabular}{c c c c c c c c c c c}
 \toprule
  KCF\(\dagger\) & MEEM\(\dagger\) & UDT\(\dagger\) & USOT\(\dagger\) & UDT-FF & TBSI$^\ast$ & ViPT$^\ast$ & AFter$^\ast$ &S2OTFormer & GDSTrack\\
  \midrule
   124.1 & 4.9 & 74.3& 58.4 & 71.9 & 50.4 &92.3 & 27.3 & 38.2 & 37.6\\
 \bottomrule
\end{tabular}
}
\caption{Speed comparison (fps) between GDSTrack and other self-supervised RGB-T tracking methods.}
\label{fps}
\end{table}
\begin{figure}[!t]
\centering
\includegraphics[width=1\linewidth]{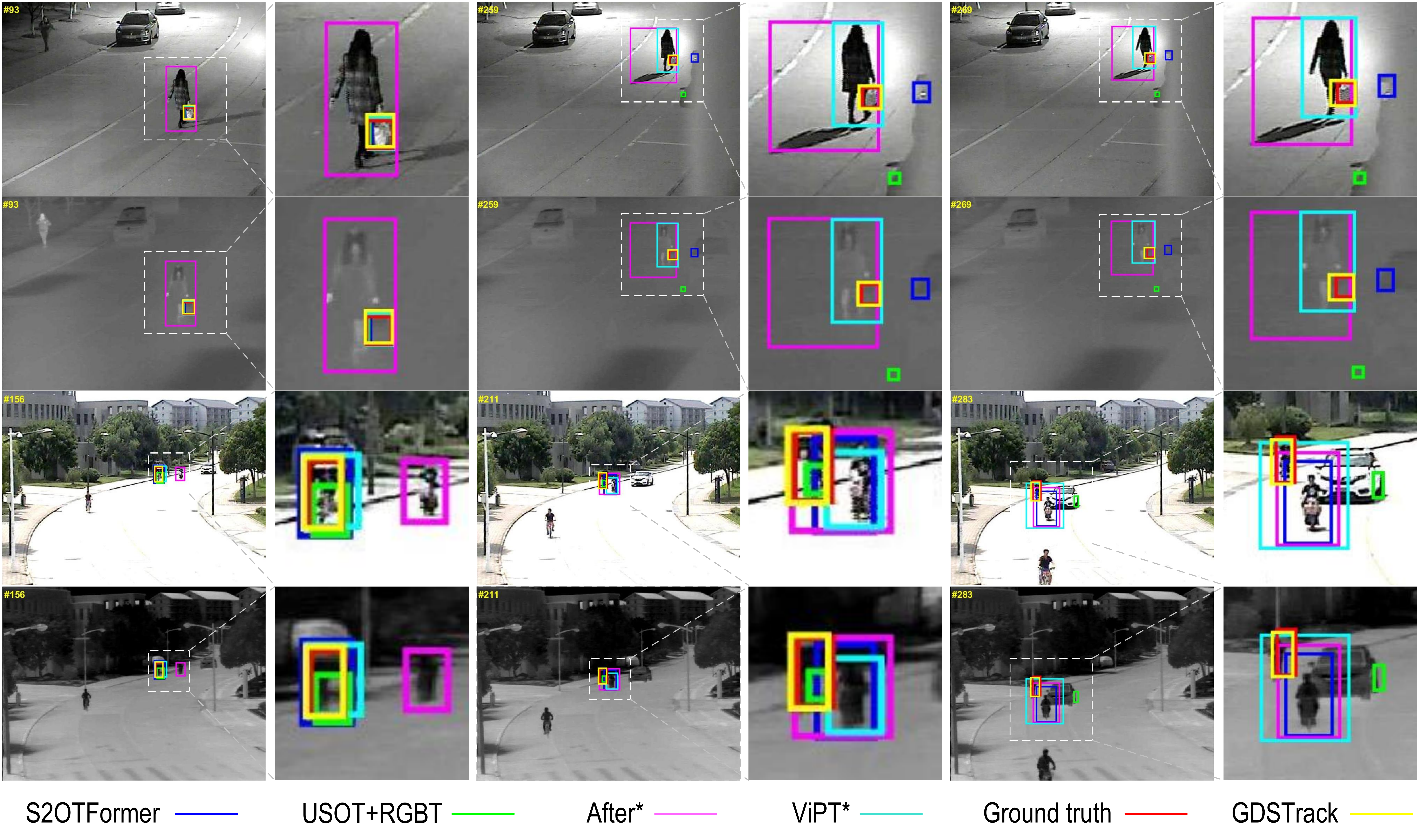}
\caption{Visualization results of GDSTrack alongside other self-supervised state-of-the-art methods on the RGBT234 dataset. The first two rows are the RGB and infrared frames of sequence \emph{baginhand}, and the last two rows are the RGB and infrared frames of sequence \emph{cycle2}. For complete sequence visualization details, please refer to the appendix.}
\label{RGBT234-visualize-zoomin}
\end{figure}

\noindent \textbf{Visualization Analysis.} We conducted visualization on the RGBT234 dataset. As shown in Fig.~\ref{RGBT234-visualize-zoomin}, for better visualization, we magnify the objects within the dashed boxes. Through comparison, we can observe that, thanks to our MDGF module, our method can locate the object area more accurately. For example, in sequence \emph{baginhand}, as the object is tracked, the ViPT$^\ast$ method gradually focuses on the pedestrian, while the After$^\ast$ method simultaneously tracks the shadow region. In contrast, our method precisely tracks the pedestrian's bag in hand. Furthermore, benefiting from our TGID module, our method demonstrates stronger resistance to similar object noise. For instance, in sequence \emph{cycle2}, when tracking the pedestrian, other methods gradually get distracted by similar objects and begin tracking a different pedestrian with similar features, whereas our method consistently tracks the same pedestrian.
\subsection{Ablation Study}
\noindent \textbf{Component Analysis.} We conducted ablation studies on four datasets to evaluate the effectiveness of the MDGF and TGID modules, with the results presented in Tab.~\ref{Ab-components}. We constructed our baseline model by directly incorporating RGB and infrared features into the RGB tracker, using a model trained with pseudo-labels. The baseline results on the four datasets are presented in the first row of Tab.~\ref{Ab-components}.
\begin{table}[t]
\centering
\resizebox{0.5\textwidth}{!}{
\begin{tabular}{c c c c c c} 
 \toprule
     MDGF & TGID & RGBT234 & LasHeR & VTUAV & GTOT\\
   \midrule
   & & 69.1/47.1 & 43.8/37.6/33.7 & 69.0/56.2 &68.6/56.0\\
   \checkmark & & 70.9/48.3 & 45.9/39.1/35.2& 70.3/58.0& 73.5/59.5 \\
   \checkmark & \checkmark& $\bm{70.9}$/$\bm{48.5}$ & $\bm{45.9}$/$\bm{39.3}$/$\bm{35.4}$&$\bm{72.3}$/$\bm{59.8}$& $\bm{73.9}$/$\bm{59.8}$ \\
 \bottomrule
\end{tabular}
}
\caption{PR, NPR, and SR of the model with or without MDGF or TGID, evaluated on four datasets.}
\label{Ab-components}
\end{table}
\begin{table}[t]
\centering
\resizebox{0.5\textwidth}{!}{
\begin{tabular}{c c c c c} 
 \toprule
  Adj. Matrix & RGBT234 & LasHeR & VTUAV & GTOT \\
   \midrule
   Identity & 69.8/$\bm{48.7}$ & 42.3/37.5/33.1 & 68.7/57.3& 70.8/58.3\\
   QKV & 67.6/46.2 & 42.5/36.5/32.7 & 70.0/57.6& 71.9/57.5 \\
   Cosine & 69.2/46.4 & 42.8/36.4/33.0 & 68.7/56.5& 70.6/58.6 \\
   AMG (Ours) & $\bm{70.9}$/48.3 & $\bm{45.9}$/$\bm{39.1}$/$\bm{35.2}$& $\bm{70.3}$/$\bm{58.0}$& $\bm{73.5}$/$\bm{59.5}$ \\
 \bottomrule
\end{tabular}
}
\caption{PR, NPR, and SR of the model with different adjacency matrix generation methods, evaluated on four datasets.}
\vspace{-3mm}
\label{Ab-stage1}
\end{table}

\noindent \textbf{The ablation of MDGF.} The tracker with MDGF demonstrates a notable improvement over the baseline model, with PR scores of 1.8\%, 1.2\% on the RGBT234dataset, 2.1\%, 1.5\%, and 1.5\% on the LasHeR dataset, 1.3\% and 1.8\% on the VTUAV dataset, and 4.9\% and 3.5\% on the GTOT dataset. As shown in the first and second rows of this table, our MDGF module can dynamically adjust the adjacency matrix, thereby better integrating the advantages of both modalities for subsequent tracking.
\begin{table}[t]
\centering
\resizebox{0.5\textwidth}{!}{
\begin{tabular}{c c c c c c c}
 \toprule
  $k$ & 5 & 10 & 15 & 20 & 128 & 256\\
  \midrule
   GTOT & 69.9/57.0 & 70.6/56.6 & 69.5/54.2&69.5/54.7&71.9/58.1&$\bm{73.5}$/$\bm{59.5}$\\
   RGBT234& 68.9/47.7 & 67.8/46.2 &69.0/46.3 &66.8/45.2&69.6/47.0&$\bm{70.9}$/$\bm{48.3}$\\
 \bottomrule
\end{tabular}
}
\caption{PR and SR of the model with different $k$ value evaluated on GTOT and RGBT234 datasets.}
\label{TopK}
\end{table}
\begin{table}[t]
\centering
\resizebox{0.5\textwidth}{!}{
\begin{tabular}{c c c c c c c} 
 \toprule
Naive DM & condition & distractor & RGBT234 & LasHeR & VTUAV & GTOT \\
   \midrule
 \checkmark &  &  & 68.5/47.5 & 44.0/37.9/34.0 &70.8/58.7& 69.0/56.5\\
 \checkmark & \checkmark &  & 68.9/47.6 &43.8/37.4/33.7 &71.3/58.9& 69.8/56.6\\
 \checkmark & \checkmark & \checkmark & $\bm{69.6}$/ $\bm{48.3}$ & $\bm{44.7}$/$\bm{38.3}$/$\bm{34.5}$&$\bm{72.3}$/$\bm{59.8}$& $\bm{70.0}$/$\bm{57.5}$ \\
 \bottomrule
\end{tabular}
}
\caption{PR, NPR, and SR of the model with or without middle conditions or distractor noise, evaluated on four datasets. To save computational time, we conducted this ablation experiment under the condition that $K$ is set to 5 in Tab.~\ref{TopK}.}
\label{Ab-diffusion}
\end{table}

\noindent \textbf{The ablation of TGID.} The tracker with both MDGF and TGID achieves further improvement over the MDGF-only model, with SR of 0.2\% on the RGBT234 dataset, NPR and SR of 0.2\%, 0.2\% on the LasHeR dataset, PR and SR of 2\% and 0.8\% on the VTUAV dataset, and 0.4\% and 0.3\% on the GTOT dataset.
From the second and third rows of this table, we can see that the TGID module further enhances the model's performance by improving robustness to interfering noise, building on the MDGF module.

\noindent \textbf{The ablation of AMG.} To evaluate the effectiveness of our proposed adjacency matrix generation method, we compared it with other generation methods, as shown in Tab.~\ref{Ab-stage1}. We employed the identity matrix, self-attention, and cosine similarity methods to generate adjacency matrices, which serve as comparative methods for our proposed AMG module. We can observe that training with the adjacency matrix generated by the AMG method yields the best performance on the LasHeR, VTUAV, and GTOT datasets. On the RGBT234 dataset, the AMG method showed a little decrease of 0.4\% in SR, compared to the identity matrix method. The excellent performance of the AMG module proves that, compared to other adjacency matrix generation methods, the AMG module can dynamically capture the inter-modal similarity, direct the model to focus on the object-related regions through graph attention and obtain fusion results that are favorable for tracking.

\noindent \textbf{The ablation of TopK.} Tab.~\ref{TopK} shows the performance of different values of $K$ in the AMG module on two datasets. We observe that the model's performance is positively correlated with the value of TopK. When $K$ is set to 256, the model performs the best on GTOT and RGBT234 datasets. 

\noindent \textbf{The ablation of Graph-informed condition.} As shown in Tab.~\ref{Ab-diffusion}, compared to ~\cite{DDIM}, using the first layer of graph convolution from the MDGF module as a condition for the intermediate layers of the UNet in the diffusion model resulted in a clear improvement on three datasets.

\noindent \textbf{The ablation of distractor noise.} As shown in Tab.~\ref{Ab-diffusion}, adding similar distractor noise resulted in increases of 0.7\% in PR and SR on the RGBT234 dataset, 1\% and 0.9\% on the VTUAV dataset, and 0.2\% and 0.9\% on the GTOT dataset.
\section{Conclusion}
In conclusion, we have designed a Dynamic Graph Diffusion framework to address the issues in self-supervised RGB-T tracking tasks. Specifically, to tackle the problem of incorrect fusion of non-object regions, the MDGF module has been proposed to dynamically adjust the adjacency matrix based on the similarity between modalities, guiding graph attention to focus on the fusion of coherent object regions. To address the noise interference caused by similar objects, the TGID module has incorporated the MDGF fusion results from neighboring frames as noise, training the model to improve its robustness against interference from similar objects. Experiments on four datasets have demonstrated that GDSTrack achieves state-of-the-art results in the field of self-supervised RGB-T tracking, validating the effectiveness of our model.  

\section*{Acknowledgments}

This work was supported in part by the National Natural Science Foundation of China under Grant Nos. 62172417, 62272461, 62276266 and 62472424.

\bibliographystyle{named}
\bibliography{ijcai25}

\end{document}


\appendix{

  \begin{center}
    {\bf \Large Appendix}
  \end{center}

To provide a more accurate description of the proposed model, GDSTrack, we have included some supplementary experiments and visualization details in this section.
\begin{itemize}
    \item $\lambda$: The weight of the diffusion loss in the loss function.
    \item $\beta$: The weight of the distractor noise in the total noise.
    \item $nhead$: the number of attention heads in MDGF.
    \item $PR$ plots: The precision rate under different thresholds.
    \item $NPR$ plots: The normalized precision rate under different thresholds.
    \item $SR$ plots: The success rate under different thresholds.
    \item Analysis of whether distractor noise requires background filtering.
    \item More attribute visualization on VTUAV dataset.
    \item More visualization of the tracking results.
\end{itemize}
\section{Analysis of $\lambda$} 
\begin{table}[h]
\centering
\resizebox{0.4\textwidth}{!}{
\begin{tabular}{c c c c c}
 \toprule
  $\lambda$ & 3 & 5 & 7 & 10\\
  \midrule
   GTOT & 69.6/56.3 & 70.0/$\bm{57.5}$ & 69.6/56.8 & $\bm{70.2}$/56.9\\
   RGBT234 & 68.6/47.6 & $\bm{69.6}$/$\bm{48.3}$ & 68.4/47.6 & 68.5/47.6\\
 \bottomrule
\end{tabular}}
\caption{PR/SR of GDSTrack with different contribution of generative loss when $\beta$ is 0.5.}
\label{lambda}
\end{table}
We use $\lambda$ in Equation (9) to balance the tracking process and the diffusion process, as shown in Tab.~\ref{lambda}. A larger lambda indicates a higher weight for the diffusion loss. We conduct experiments with $\lambda$ set to (3, 5, 7, 10). We find that when $\lambda$ is set to 5, the model performs best on the GTOT and RGBT234 benchmarks. When $\lambda$ exceeds 5, the model starts to decline due to the tracking task's weight being too small. We found that even with a smaller $\lambda$, our model outperforms most state-of-the-art methods. Tab.~\ref{lambda} analyzes the weights of the diffusion model in Equation (9). We experimented with the condition that $K$ is set to 5.
\section{Analysis of $\beta$} We analyze the weight of the distractor noise in Equation (20) in Tab.~\ref{beta}. The larger the $\beta$, the greater the proportion of distractor noise, and the smaller the proportion of Gaussian noise. When $\beta$ is set to 0, it means no distractor noise is added. We found that setting the noise value to 0.1 yields better performance than using only Gaussian noise. The best performance is achieved when the noise ratio is set to 0.5. When the noise ratio exceeds 0.5, the model's performance starts to degrade due to the introduction of excessive noise. We ultimately choose to add noise when a random number $p > 0.5$, with $\beta$ set to 0.5, to enhance the model's robustness while avoiding the introduction of excessive noise. To save computational time, we experimented with the condition that $K$ is set to 5.
\begin{table}[h]
\centering
\resizebox{0.5\textwidth}{!}{
\begin{tabular}{c c c c c c c c}
 \toprule
  $\beta$ & 0 & 0.1 & 0.2 & 0.3 &0.4 ($p$)& 0.5 ($p$)&0.6 ($p$)\\
  \midrule
   GTOT & 69.8/56.6 & $\bm{70.3}$/57.4 & 68.5/56.3 & 69.2/56.3 &70.1/57.3& 70.0/$\bm{57.5}$& 69.4/57.0\\
   RGBT234 & 68.9/47.6 & 69.3/48.2 & 68.4/47.7 & 69.1/47.8 &69.2/48.1& $\bm{69.6}$/$\bm{48.3}$&69.5/48.1\\
 \bottomrule
\end{tabular}
}
\caption{PR/SR of GDSTrack with different contribution of distractor noise when $\lambda$ is 5. $p$ indicates the use of distractor noise when the random number $p > 0.5$.}
\label{beta}
\end{table}
\section{Analysis of $nhead$}
\begin{table}[h]
\centering
\resizebox{0.3\textwidth}{!}{
\begin{tabular}{c c c c}
 \toprule
  $nhead$ &1& 2 & 3 \\
  \midrule
   GTOT & 69.9/57.0 &70.1/55.9& $\bm{71.0}$/$\bm{57.2}$ \\
   RGBT234 & $\bm{68.9}$/$\bm{47.7}$ &67.1/46.8& 64.8/45.2 \\
 \bottomrule
\end{tabular}
}
\caption{PR/SR of the model with different $nhead$ value of GAT evaluated on GTOT and RGBT234 datasets when $k$ in AMG is 5.}
\label{nhead}
\end{table}

\begin{table}[h]
\centering
\resizebox{0.3\textwidth}{!}{
\begin{tabular}{c c c}
 \toprule
  distractor noise & whole feature & filter background\\
  \midrule
   GTOT & $\bm{70.0}$/$\bm{57.5}$ & 69.6/56.6\\
   RGBT234 & $\bm{69.6}$/$\bm{48.3}$ & 69.2/47.8\\
 \bottomrule
\end{tabular}}
\caption{PR/SR scores of GDSTrack when using complete features as distractor noise versus utilizing the first-stage tracking results to retain the object parts in the features.}
\label{filt-bg}
\end{table}

We analyzed the $nhead$ parameter of the graph attention in MDGF, as shown in Tab.~\ref{nhead}. $nhead$ indicates the number of attention heads in MDGF. We did not find a positive correlation between the model's performance and $nhead$. In addition, a larger value of $n$ leads to higher computational complexity and increased time costs. Therefore, we set $nhead$ to 1.
\section{Analysis of $PR$, $NPR$, and $SR$ plots}
\begin{figure}[!ht]
\centering
\includegraphics[width=3in]{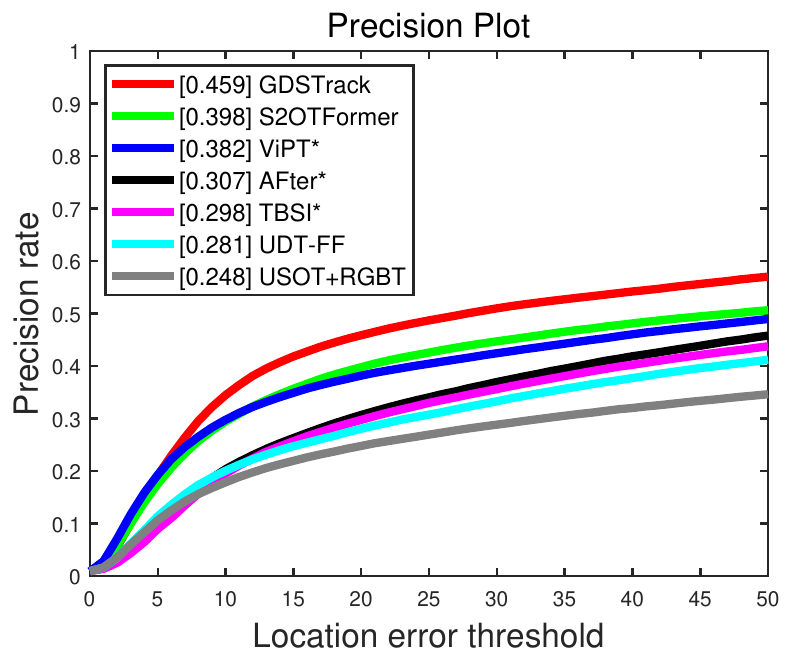}%
\caption{PR on LasHeR dataset compared against other self-supervised RGBT trackers.}
\label{LasHeR-PR}
\end{figure}
\begin{figure}[!ht]
\centering
\includegraphics[width=3in]{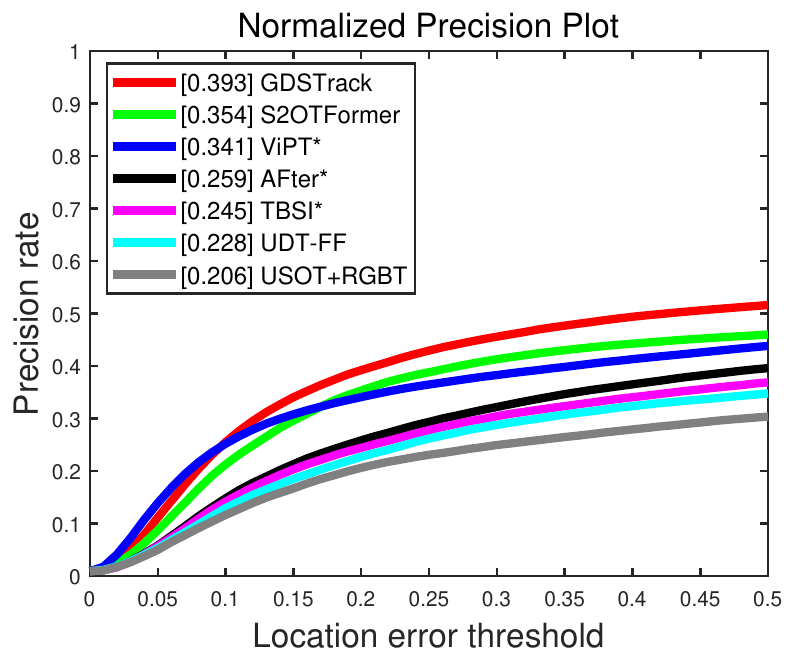}%
\caption{NPR on LasHeR dataset compared against other self-supervised RGBT trackers.}
\label{LasHeR-NPR}
\end{figure}

\begin{figure}[!ht]
\centering
\includegraphics[width=3in]{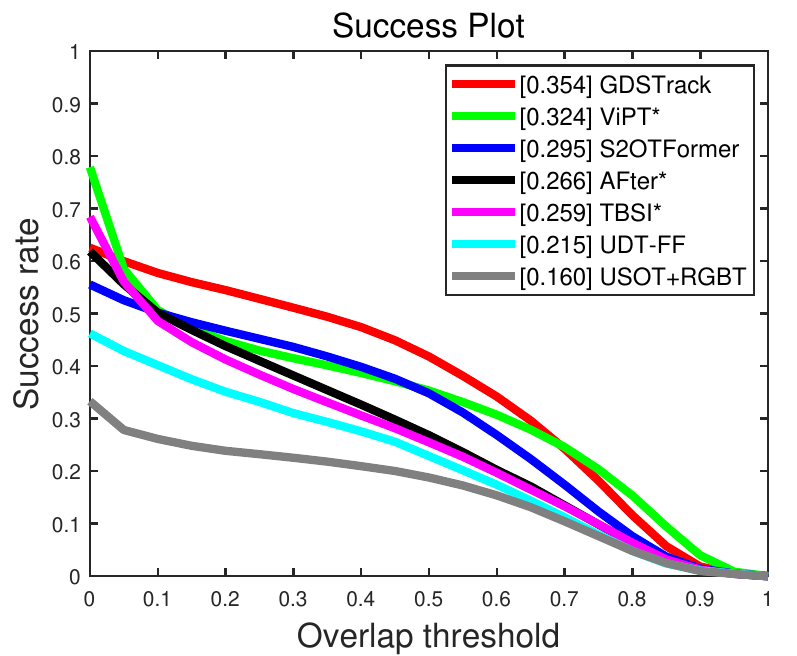}%
\caption{SR on LasHeR dataset compared against other self-supervised RGBT trackers.}
\label{LasHeR-SR}
\end{figure}
We compared our model with other state-of-the-art self-supervised RGB-T trackers on the LasHeR dataset. As shown in Fig.~\ref{LasHeR-PR}, Fig.~\ref{LasHeR-NPR}, and Fig.~\ref{LasHeR-SR}, our model achieved the best performance on both the PR plots, NPR plots, and the SR plots.

We compared our model with other state-of-the-art self-supervised RGB-T trackers on the VTUAV dataset. As shown in Fig.~\ref{VTUAV-PR} and Fig.~\ref{VTUAV-SR}, our model achieved the best performance on both the PR plots and the SR plots.
\begin{figure}[!ht]
\centering
\includegraphics[width=3in]{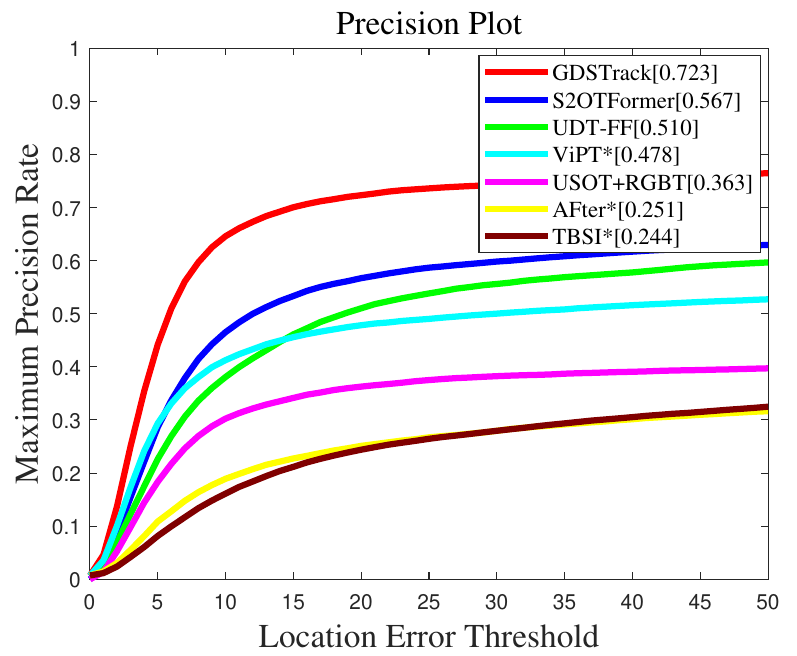}%
\caption{PR on VTUAV dataset compared against other self-supervised RGBT trackers.}
\label{VTUAV-PR}
\end{figure}

\begin{figure}[!ht]
\centering
\includegraphics[width=3in]{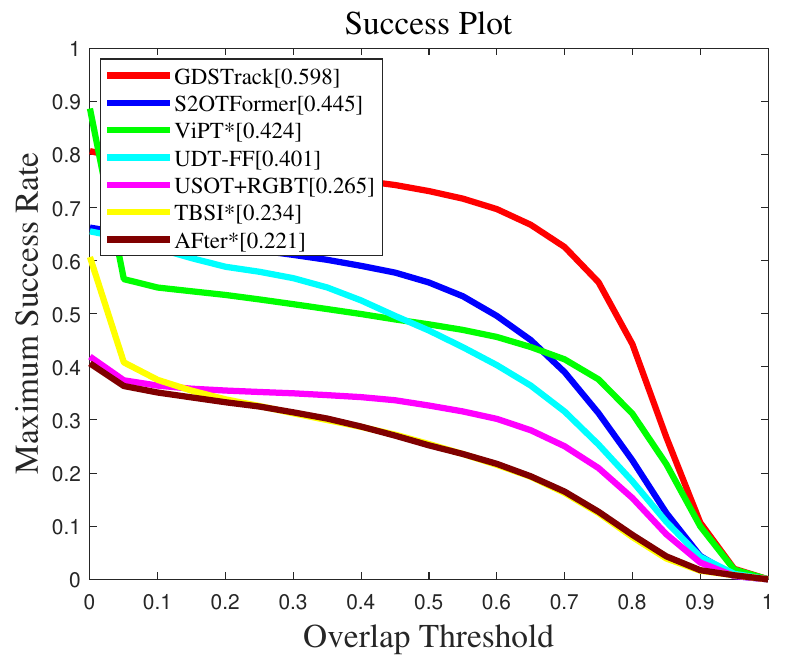}%
\caption{SR on VTUAV dataset compared against other self-supervised RGBT trackers.}
\label{VTUAV-SR}
\end{figure}

\begin{figure}[h]
\centering
\subfloat[]{\includegraphics[width=1.7in]{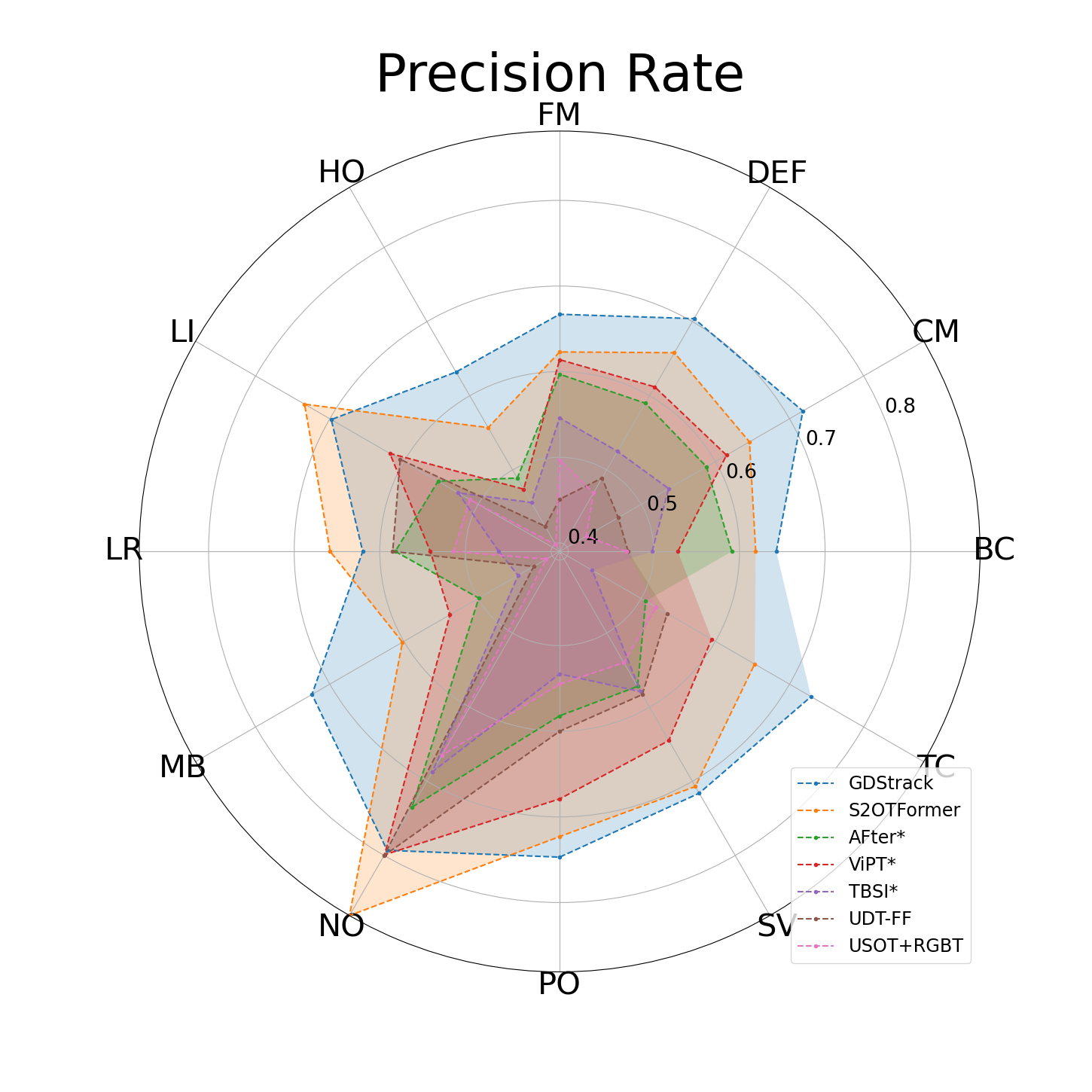}}%
\subfloat[]{\includegraphics[width=1.7in]{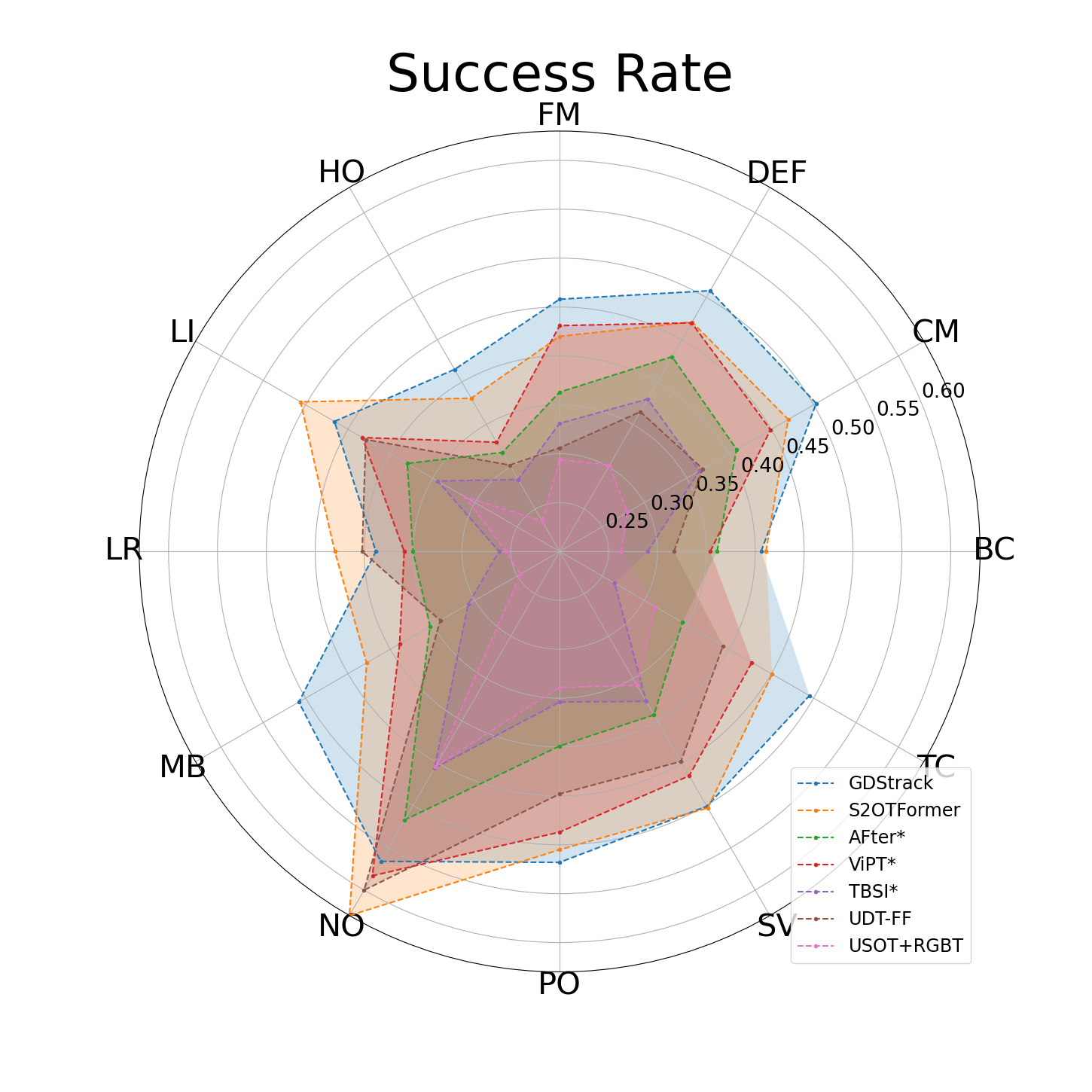}}%
\caption{Attribute-based evaluation on RGBT234 dataset compared against five self-supervised RGBT trackers. (a) Precision Rate with different attributes. (b) Success Rate with different attributes.}
\label{RGBT234-attr}
\end{figure}

\begin{figure}[!ht]
\centering
\includegraphics[width=3in]{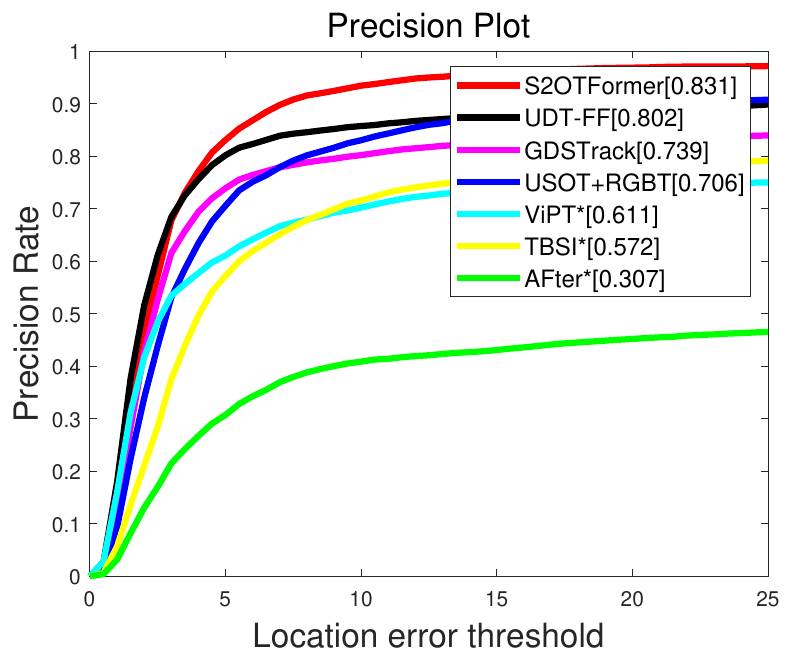}%
\caption{PR on GTOT dataset compared against other self-supervised RGBT trackers.}
\label{GTOT-PR}
\end{figure}

\begin{figure}[!ht]
\centering
\includegraphics[width=3in]{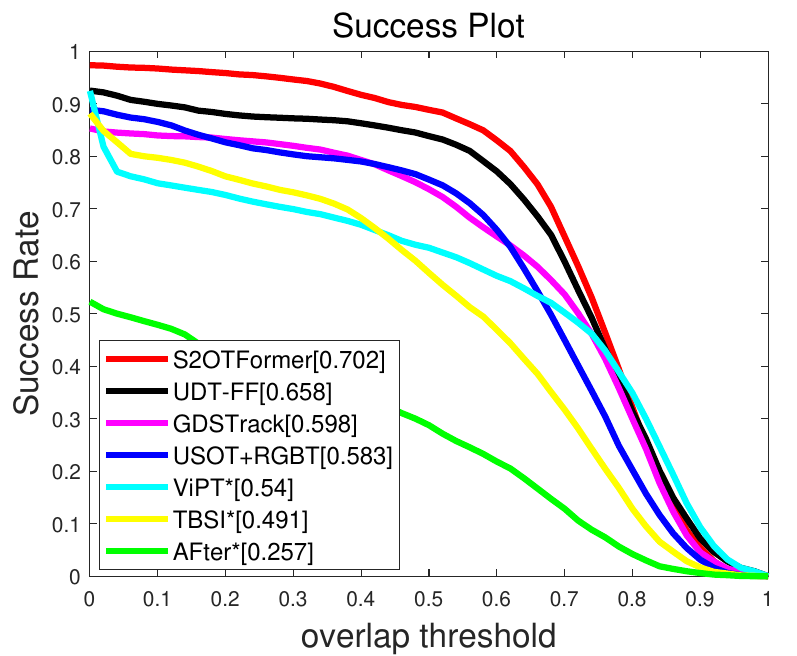}%
\caption{SR on GTOT dataset compared against other self-supervised RGBT trackers.}
\label{GTOT-SR}
\end{figure}

\begin{figure}[!t]
\centering
\includegraphics[width=1\linewidth]{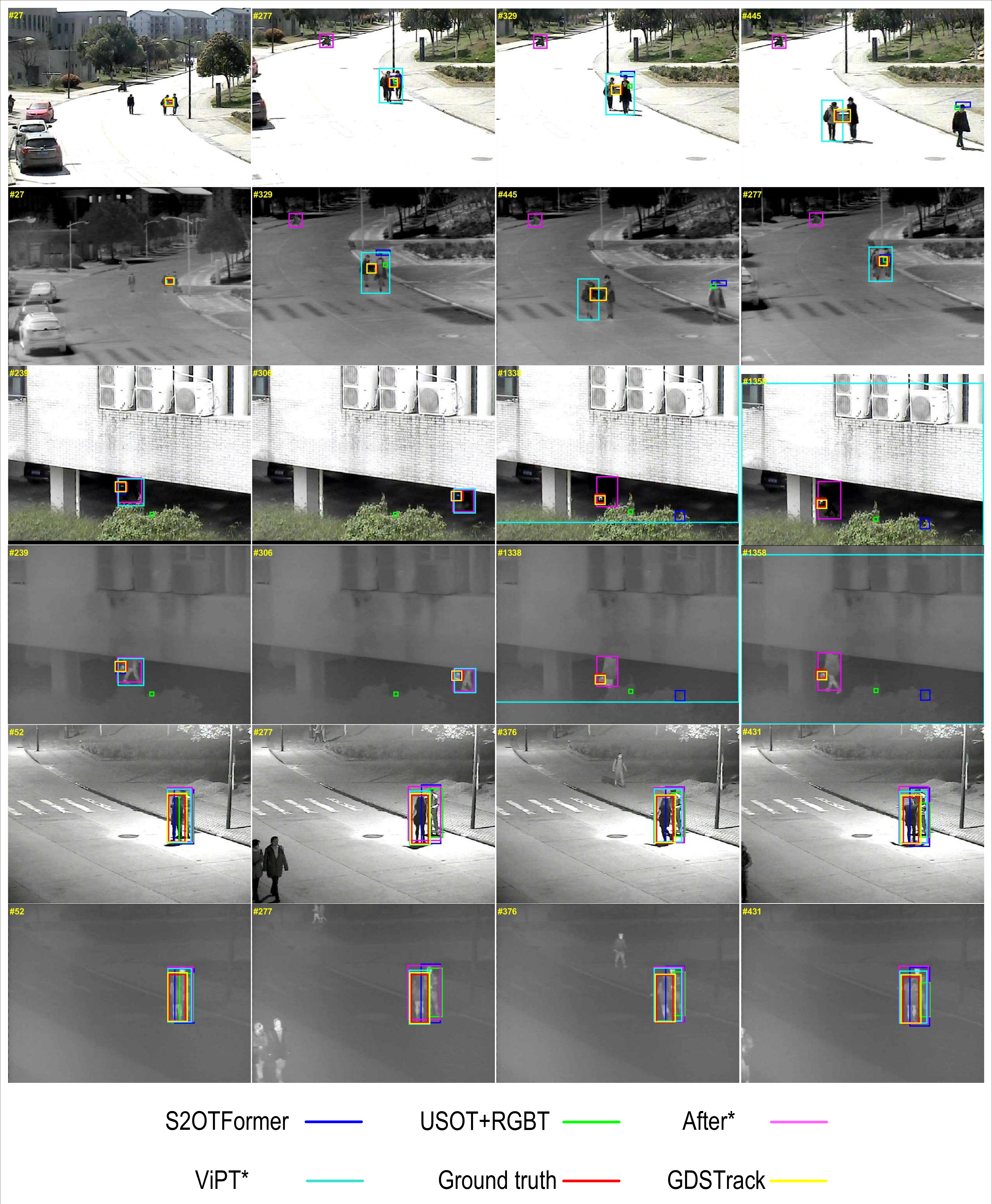}
\caption{Visualization results of GDSTrack alongside other self-supervised state-of-the-art methods on \emph{man26}, \emph{kettle}, and \emph{jump} sequence in the RGBT234 dataset.}
\label{RGBT234-visualize}
\end{figure}
We compared our model with other state-of-the-art self-supervised RGB-T trackers on the GTOT dataset, as shown in Fig.~\ref{GTOT-PR} and Fig.~\ref{GTOT-SR}. Our model did not achieve the best performance on the GTOT dataset. We hypothesize that this is due to the smaller number of sequences and shorter sequence lengths in the GTOT dataset compared to the other three datasets, particularly when compared to the much larger VTUAV dataset.
\section{Analysis of background filtering.}
When introducing distractor noise, we use the tracking results from MDGF to filter out the background regions where noise is introduced. From Tab.~\ref{filt-bg}, we can observe that removing the background leads to a decrease in the model's performance. We hypothesize that this is because the background itself contains valuable contextual information, and removing it may result in the loss of key information. We conducted this ablation experiment under the condition that $K$ is set to 5.
\section{Analysis of Attribute.} The performance of GDSTrack, along with other sota self-supervised methods on different attribute challenges is shown in Fig.~\ref{RGBT234-attr}. As observed, our method achieves the best performance on ten attributes with 7.6\% of PR and 4.4\% of SR higher than S2OTFormer on the Thermal Crossover attribute. Our method achieves 8.8\% of PR and 6.4\% of SR lower than S2OTFormer on the No Occlusion attribute. Our method achieves 3.6\% of PR and 4\% of SR lower than S2OTFormer on the Low Illumination attribute.
\section{More visualization of the tracking results}
We present a more comprehensive visualization of the model's sequence results on the RGBT234 dataset in Fig.~\ref{RGBT234-visualize}. The visualization consists of six rows, where the first row represents the RGB modality and the second row represents the infrared modality. This pattern is repeated for three sequences, \emph{man26}, \emph{kettle}, and \emph{jump} sequences. Compared to other sota self-supervised trackers, we observed that our method exhibits better interference resistance for similar objects. Specifically, we have indeed addressed the issues of pseudo-label background noise and interference from similar objects. Our method alleviates the problem of tracking large objects that move together with small objects during the tracking process. It also reduces the interference caused by similar objects, such as nearby pedestrians, to the object.
}